\pdfoutput=1

\documentclass[11pt]{article}

\usepackage[preprint]{acl}

\usepackage{times}
\usepackage{latexsym}

\usepackage[T1]{fontenc}

\usepackage[utf8]{inputenc}

\usepackage{microtype}

\usepackage{inconsolata}

\usepackage{graphicx}
\usepackage{hyperref}
\usepackage{url}
\usepackage{enumitem}
\usepackage[most]{tcolorbox}

\usepackage{booktabs}
\usepackage{booktabs}
\usepackage{caption}
\usepackage{wrapfig}
\usepackage{CJKutf8}
\usepackage{multirow}
\usepackage{colortbl}
\usepackage{amsfonts} 
\usepackage{fontawesome}
\usepackage[normalem]{ulem}
\useunder{\uline}{\ul}{}

\title{LongReward: Improving Long-context Large Language Models\\ with AI Feedback
}

\author{%
  Jiajie Zhang$^{1\dagger}$, Zhongni Hou$^{2\dagger}$, Xin Lv$^{3}$, Shulin Cao$^{3}$, Zhenyu Hou$^1$, Yilin Niu$^3$,\\ 
  \textbf{Lei Hou$^1$, Yuxiao Dong$^1$, Ling Feng$^1$, Juanzi Li$^1$} \\
  $^1$Tsinghua University
  \quad
  $^2$University of Chinese Academy of Sciences
  \quad
  $^3$Zhipu AI
}

\begin{document}
\maketitle

\renewcommand{\thefootnote}{\fnsymbol{footnote}}
    \footnotetext[2]{Work done when JZ and ZH interned at Zhipu.AI.
    }

\begin{abstract}
Though significant advancements have been achieved in developing long-context large language models (LLMs), the compromised quality of LLM-synthesized data for supervised fine-tuning (SFT) often affects the long-context performance of SFT models and leads to inherent limitations. In principle, reinforcement learning (RL) with appropriate reward signals can further enhance models' capacities. However, how to obtain reliable rewards in long-context scenarios remains unexplored. To this end, we propose \textbf{LongReward}, a novel method that utilizes an off-the-shelf LLM to provide rewards for long-context model responses from four human-valued dimensions: helpfulness, logicality, faithfulness, and completeness, each with a carefully designed assessment pipeline. By combining LongReward and offline RL algorithm DPO, we are able to effectively improve long-context SFT models. Our experiments indicate that LongReward not only significantly improves models' long-context performance but also enhances their ability to follow short instructions. We also find that long-context DPO with LongReward and conventional short-context DPO can be used together without hurting either one's performance. Our code and data are available at \url{https://github.com/THUDM/LongReward}.
\end{abstract}

\section{Introduction}
\label{sec:intro}
In recent years, significant advancements have been achieved in the development of long-context large language models (LLMs)~\cite{claude-35, glm4, reid2024gemini}.  The context windows of many contemporary LLMs have been extended to over 100,000 tokens, enabling them to process extensive context as input and perform various downstream tasks such as long document understanding and summarization~\cite{longbench}. 

Among numerous methods for extending the context window, an effective and well-established approach involves continual pre-training on longer texts, followed by supervised fine-tuning (SFT) using diverse long-context question-answering (QA) data~\cite{longllama, longalign}.  However, due to the difficulty of annotation, most long-context QA pairs are automatically synthesized by LLMs themselves~\cite{longalign, llama-3-1, An2024}, making it challenging to guarantee the quality of data. For instance, the generated answers may not fully adhere to query requirements,  contain logical errors, include fabricated content, or be incomplete. Such compromised data quality often affects the long-context capacities of SFT models, making them suffer from inherent flaws such as hallucinations~\cite{huang23} and inability to fully utilize the context information~\cite{lostinmiddle, longcite}.

On the other hand, reinforcement learning (RL) with human- or AI-based rewards that penalize unpreferred behaviors has been shown as effective in reducing irrelevant, untruthful, and other undesired model outputs for short-context LLMs~\cite{rlhf, rlaif, tian2024}. Nevertheless, obtaining such rewards in long-context scenarios is still challenging due to the unscalability of human labeling and the lack of reliable long-context reward models. 

\begin{figure*}
    \centering
    \includegraphics[width=\linewidth]{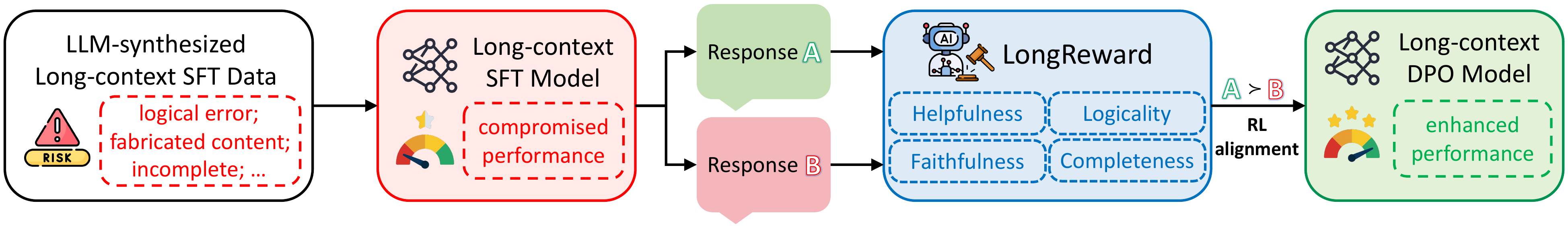}
    \caption{The compromised quality of synthesized SFT data often affects the performance of long-context SFT models, while LongReward utilizes an off-the-shelf LLM to provide reliable rewards for long-context-based model responses, enabling the employment of RL algorithms such as DPO to further enhance models' capacities.}
    \label{fig:intro}
\end{figure*}

In light of these challenges, we propose \textbf{LongReward}, a novel method that utilizes an off-the-shelf LLM as judge to provide rewards for long-context-based model responses from four human-valued dimensions: (1) Helpfulness: whether the response is relevant and informative to the query and meets all the requirements; (2) Logicality: whether different parts of the response are logically consistent; (3) Faithfulness: whether all information in the response is consistent with the context; (4) Completeness: whether the response covers all question-relevant key points in the context, without omitting important aspects. Given a model response, LongReward will give a score ranging from 0 to 10 for each dimension, and take their average as the final reward. Specifically, for helpfulness and logicality whose assessment primarily depends on the response content and is mostly independent of the context, we employ the LLM to directly assign scores based on the query and response through few-shot learning. For the estimation of faithfulness, we require the LLM to first break the response into a list of factual statements and then judge whether each statement is supported by the retrieved context chunks. Finally, for completeness, we first let the LLM extract question-relevant information from each segment of the context, then ask it again to evaluate the response completeness according to all the extracted information.



By combining LongReward and RL algorithms such as Direct Preference Optimization (DPO)~\cite{dpo}, we can effectively mitigate the deficiencies of long-context SFT models and further enhance their capabilities. Our experiments on Llama-3.1-8B~\cite{llama-3-1} and GLM-4-9B~\cite{glm4} show that the DPO models using LongReward outperform SFT models by 4.9\% and 5.5\% on long-context tasks, respectively, surpassing all baseline methods. Human evaluation further validates that LongReward has good alignment with human preference and helps improve long-context models from all dimensions (i.e., helpfulness, logicality, faithfulness, and completeness), bringing 46\% more wins against the SFT baseline. Meanwhile, we find that LongReward also benefits models' short-instruction-following ability, and can be well incorporated into standard short-context DPO to jointly improve long- and short-context performance.

To summarize, our main contributions include: (1) proposing LongReward, the first method as we know to automatically provide reliable rewards for long-context-based model responses; (2) designing a long-context RL framework by combining LongReward and DPO; (3) conducting extensive experiments to validate the efficacy of LongReward in improving long-context LLMs.

\section{Related Work}
\textbf{Long-context LLMs}. 
Long-context LLMs aim to break the context length limitations of existing LLMs and understand internal long-range dynamics~\cite{longalign,ma2024megalodon}. One research direction focuses on designing efficient attention mechanisms~\cite{longformer, bigbird, minference} or structured state space models~\cite{poli2023hyena,gu2023mamba} to overcome the length limitations. For instance, ~\citet{ding2023longnet} adopts sparse attention to scale the context length to billions of tokens. However, the performance of these methods usually falls short of standard Transformers~\cite{gu2023mamba,ma2024megalodon}.
Another research branch focuses on extending Transformers' context window via continual pre-training and SFT on longer texts~\cite{longllama,longalign}. Despite larger computation overhead, these methods typically demonstrate better performance on various long-context tasks. Nevertheless, their use of automatically synthesized SFT data that lacks human examination still compromises the capacities of current long-context LLMs to some extent.

\noindent\textbf{Improving LLMs with AI feedback.} 
Reinforcement learning from human feedback is crucial in aligning LLMs with human values and intentions~\cite{rlhf, bai2022training,sun2023salmon}. However, collecting high-quality human pairwise preference data can be expensive and time-consuming~\cite{constitutionalAI, rlaif}. 
An alternative solution is to obtain feedback from LLMs, as modern LLMs have shown a high degree of alignment with human judgment~\cite{ding2022gpt,gilardi2023chatgpt}. Following this direction, ~\citet{constitutionalAI} first integrates LLM-labeled preferences data with human-labeled ones to optimize models' harmlessness.
~\citet{dubois2024alpacafarm} further introduces the AlpacaFarm simulator, which leverages API LLMs to choose the preferred model responses, offering lower costs compared to human labelers. 
More recently, ~\citet{yuan2024self} develops self-rewarding language models, wherein the LLM itself acts as a judge, providing its rewards during training. However, these diverse approaches mainly focus on short-context scenarios. In contrast, our work first incorporates AI feedback with long-context scenarios and successfully improves LLMs' long-context capability.

\begin{figure*}
    \centering
    \includegraphics[width=0.95\linewidth]{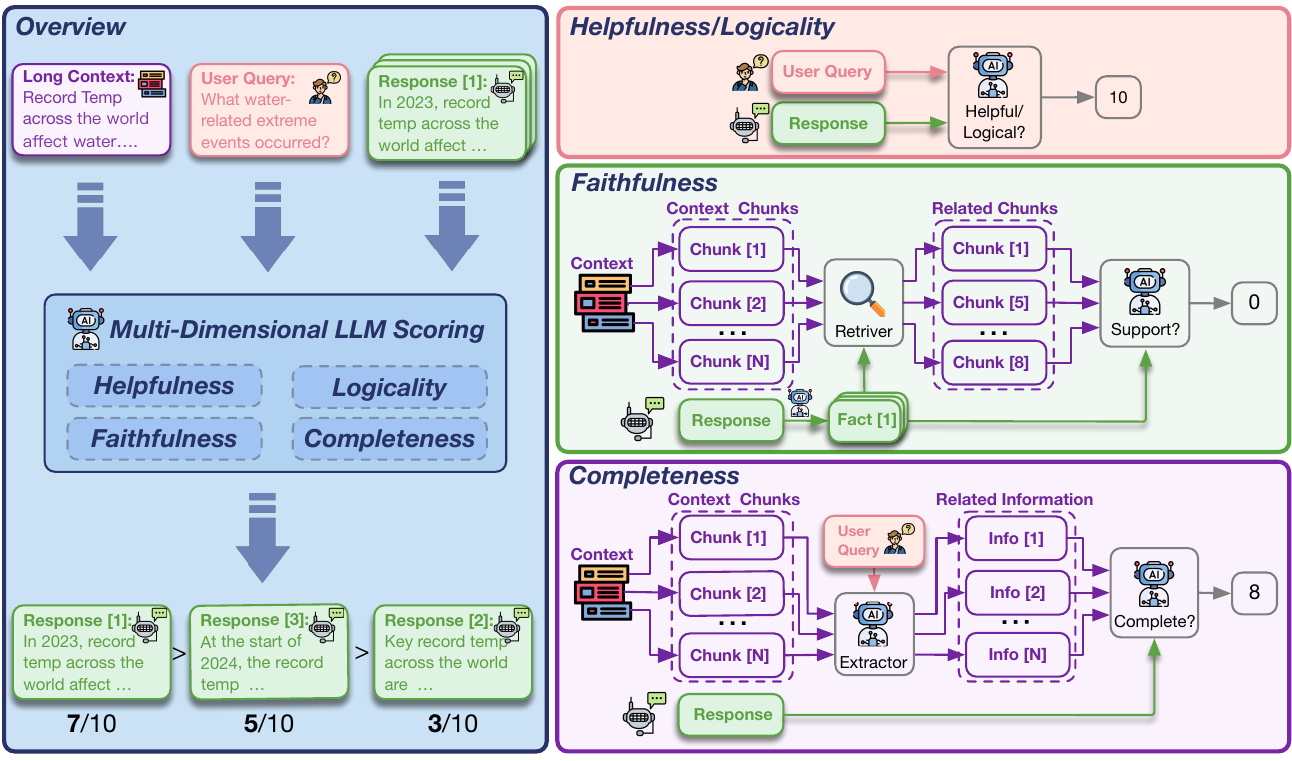}
    \caption{Illustration of LongReward. LongReward evaluates a long-context-based model response from four dimensions: helpfulness, logicality, faithfulness, and completeness. It assigns a score ranging from 0 to 10 for each dimension, and takes their average as the final reward.}
    \label{fig:longreward}
\end{figure*}

\section{Methodology}

In this section, we will briefly introduce reinforcement learning for LLMs as well as the DPO algorithm, and then discuss the methodology of LongReward, including multi-dimensional LLM scoring for long-context-based model responses and the combination of LongReward and DPO.


\subsection{Preliminary}
Reinforcement learning (RL) aligns LLMs with human preference by maximizing the average reward of model outputs, where a reward model $r(x,y)$ assigns a scalar reward to each input-output pair $(x, y)$ to represent its desirability~\cite{rlhf, bai2022training, stiennon2020}. Conventional RL algorithms such as PPO~\cite{ppo} involve online response sampling and training multiple LLMs, thereby being complex to implement, while DPO~\cite{dpo} simplifies the RL process and proposes to directly learn from a dataset of preference pairs $\mathcal{D}=\{(x, y_w, y_l)\}$, where the winning response $y_w$ is preferred over the losing response $y_l$ given the same prompt $x$. The optimization objective of DPO is to maximize the difference between likelihood of preference pairs:  
\begin{equation}
\begin{aligned}
&\mathcal{L}_\text{DPO}(\pi_\theta; \pi_\text{ref})= -\mathbb{E}_{(x, y_w, y_l)\sim\mathcal{D}} \\
&[\log\sigma(\beta\log\frac{\pi_\theta(y_w|x)}{\pi_\text{ref}(y_w|x)}-\beta\log\frac{\pi_\theta(y_l|x)}{\pi_\text{ref}(y_l|x)})]
\end{aligned}
\end{equation}
Here, $\pi_\theta$ denotes the policy model, which is the LLM being trained and usually initialized from its SFT version, $\pi_\text{ref}$ denotes the reference model, typically the frozen SFT model, and $\beta$ is a coefficient that controls the penalty intensity for dispreferred responses. Though DPO eliminates the need for an explicit reward model, many works still train a reward model~\cite{rso, rs-dpo, chatglm-rlhf} or design proxy reward methods~\cite{tian2024} to enable automated annotations of preference pairs and efficient sampling from the SFT policy, especially when human preference labeling is costly and unscalable.


\subsection{LongReward}
As mentioned in Sec.~\ref{sec:intro}, the main obstacle to employing RL for long-context LLMs is the lack of approaches to obtain reliable rewards in long-context scenarios. 
Inspired by LLM-as-Judge approach in open-ended question-answering evaluation~\cite{mt-bench, alpacaeval2, alignbench}, we propose LongReward, a novel method that utilizes an off-the-shelf LLM $M_\text{judge}$ to provide reward signals for long-context-based model responses. As illustrated in Figure~\ref{fig:longreward}, given a long-context prompt $x$ (typically consisting of a lengthy context $c$ and a query $q$) and a response $y$, LongReward evaluates the response based on four dimensions that are valued by humans: helpfulness, logicality, faithfulness, and completeness. For each dimension, LongReward assigns a score ranging from 0 to 10, and the average of these scores constitutes the final reward. Below, we provide a detailed description of these four dimensions and their respective assessment methods. The detailed prompts are listed in Appendix~\ref{appendix:prompts}.

\paragraph{Helpfulness.}
We define a model response as ``helpful'' if it is relevant and informative to the query, fulfilling the user's intended purpose and all the requirements. Since we treat faithfulness, i.e., the truthfulness of provided information, as a separate dimension instead of an aspect of helpfulness, the assessment of helpfulness can be directly based on the query and the response without referring to the lengthy context. Specifically, we first provide $M_\text{judge}$ with detailed scoring principles and several examples with different helpfulness scores as references, then feed the query and the response into $M_\text{judge}$ and ask it to rate helpfulness for the response. Similar to~\citet{mt-bench}, we introduce Chain-of-Thought (CoT)~\cite{cot}, i.e., requiring $M_\text{judge}$ to generate an analysis before providing the final score, to augment both the score reliability and interoperability.

\paragraph{Logicality.}
Since LLMs generate responses in an autoregressive way that lacks a rollback mechanism, it is difficult for them to fix errors once generated, even if the subsequent output and the conclusion are correct. This typically results in logical inconsistencies within their responses. Additionally, current LLMs often make mistakes in simple calculation and reasoning tasks. The focus of logicality assessment is to detect such logical errors in model responses, which is also independent of the context. Therefore, similar to the assessment of helpfulness, we utilize few-shot learning with CoT to enable $M_\text{judge}$ to first find possible logical errors in the response and then rate its logicality.

\paragraph{Faithfulness.} Faithfulness measures the proportion of factual information in a model response that aligns with the context. Since it is challenging to directly find out all unfaithful information given the substantial context length, we follow the break-and-check idea of FactScore~\cite{factscore}, requiring  $M_\text{Judge}$ to first break the model response into a list of factual statements $\mathcal{S}=\{s_1, \dots, s_n\}$, and then judge whether each statement $s_i$ is supported by the most relevant context, which are top-$k$ 128-token chunks retrieved from the context taken $s_i$ as the query. Meanwhile, we make the following changes to better adapt to long-context scenarios as well as improve efficiency: (1) Current long-context models are prone to copy sentences from the context, so we break the model response into sentence-level factual statements instead of atomic statements to improve the retrieval recall and reduce the number of fact-checking; (2) Since most responses contain several ``functional sentences'' such as ``... has the following advantages:'' and ``In summary, ...'' that do not provide factual information or are conclusions or reasoning based on the previous response content, we require $M_\text{judge}$ to omit these sentences when decomposing responses to improve assessment accuracy. (3) For each factual statement $s_i$, we consider three supporting levels: full support, partial support, and no support, and set the corresponding score $a_i$ to be 1, 0.5, and 0, respectively. The final faithfulness score is calculated as $(10\cdot \sum_{i=1}^n a_i) / n$.

\paragraph{Completeness.} The focus of completeness is to ascertain whether the response covers all question-relevant key points in the context and provides sufficient information and details to meet the user's needs. Considering LLMs are likely to ignore information in the middle or tail of the context when it is extremely long~\cite{lostinmiddle, longcite}, we adopt a divide-and-conquer strategy for completeness assessment. Specifically, we first divide the context into coarse-grained chunks with a length of 4.096 tokens and ask $M_\text{judge}$ to extract question-relevant information from each chunk. Next, we concatenate all the information and call $M_{judge}$ again to assess whether the response encompasses all important aspects without any apparent omissions, then assign a final score for completeness. The scoring process is also implemented with few-shot learning and CoT.

\subsection{Long-context DPO with LongReward}
\label{sec:longreward+dpo}
Now that we have an automated approach to calculate rewards for long-context-based model responses, we can combine it with some RL algorithms to further enhance long-context SFT models. Here, we demonstrate how to combine LongReward with DPO, one of the most popular offline RL algorithms. The key to applying DPO for a given long-context SFT model $M_\text{SFT}$ is to construct a long-context preference dataset, which can be done using LongReward and a set of long-context prompts.  
These prompts can be either from the SFT dataset or newly collected. For each prompt, we first sample $m$ candidate responses from $M_\text{SFT}$ using simple temperature sampling with temperature 1.0. For each response, we then apply LongReward to obtain its reward. Following~\cite{chatglm-rlhf}, we ultimately choose the responses with the highest and lowest reward as the preference pair for the prompt. Finally, we can fine-tune $M_\text{SFT}$ with DPO to enhance its capacities. To further stabilize DPO training, we add an additional cross-entropy (CE) loss on the winning sequences as a regularization term, similar to~\cite{pan2024}: 
\begin{equation}
\mathcal{L}_\text{CE}(\pi_\theta)= -\mathbb{E}_{(x, y_w)\sim\mathcal{D}}[\log(\pi_\theta(y_w|x))],
\end{equation}

\begin{equation}
    \mathcal{L}_\text{merge} = \mathcal{L}_\text{DPO} + \lambda \cdot\mathcal{L}_\text{CE}
\end{equation}
where $\lambda$ denotes the scaling coefficient of CE loss.
\begin{table*}[t]
\centering
\resizebox{\linewidth}{!}{
\begin{tabular}{llccccc}
\toprule
Dataset                    & Task Type             & \#Data & Avg Len & Language        & Metric          & Judge Model \\ \midrule
\multicolumn{7}{l}{\textit{\textbf{Long-context Benchmark}}}                                                               \\
LongBench-Chat             & Multi-Task            & 50         & 35,571  & English/Chinese & Point-wise Rate & GPT-4o      \\ 
\multirow{3}{*}{LongBench} & Single-Doc QA          & 750        & 8,573   & English/Chinese & Point-wise Rate & GPT-4o      \\ 
                           & Multi-Doc QA           & 800        & 1,0255  & English/Chinese & Point-wise Rate & GPT-4o      \\
                           & Summarization         & 800        & 9,210   & English/Chinese & Point-wise Rate & GPT-4o      \\ \midrule
\multicolumn{7}{l}{\textit{\textbf{Short-context Benchmark}}}                                                              \\
MT-Bench                   & Instruction Following & 80         & -       & English         & Point-wise Rate & GPT-4       \\ 
AlpacaEval2                & Instruction Following & 805        & -       & English         & LC Win Rate     & GPT-4-turbo \\ \bottomrule
\end{tabular}
}
\caption{Detailed statistics of benchmarks we used for evaluation. "Avg Len" refers to the average number of words/characters in the context of English/Chinese instances. ``LC Win Rate'' denotes length-controlled Win Rate~\cite{alpacaeval2} against GPT-4-turbo.}
\label{tab:datasets} 
\end{table*}
\begin{table*}[t]
\centering
\resizebox{0.9\linewidth}{!}{
\begin{tabular}{llccccc}
\toprule
                               &                                                &                                  & \multicolumn{3}{c}{LongBench}                                                           &                             \\ \cline{4-6}
\multirow{-2}{*}{Model}        & \multirow{-2}{*}{Method}                       & \multirow{-2}{*}{LongBench-Chat} & S-Doc QA                    & M-Doc QA                    & Summ                        & \multirow{-2}{*}{Avg}       \\ \midrule
                               & {\color[HTML]{8F959E} officially post-trained} & {\color[HTML]{8F959E} 60.2}      & {\color[HTML]{8F959E} 59.3} & {\color[HTML]{8F959E} 42.9} & {\color[HTML]{8F959E} 35.3} & {\color[HTML]{8F959E} 49.4} \\
                               & SFT                                            & 69.8                             & 66.1                        & 44.5                        & 39.6                        & 55.0                        \\
                               & DPO w/ SRM                                     & 67.4                             & 65.0                        & 49.6                        & 42.7                        & 56.2                        \\
                               & DPO w/ Contrast                                  & 70.6                             & \textbf{67.8}               & 46.2                        & 40.3                        & 56.2                        \\ 
\multirow{-5}{*}{Llama-3.1-8B} & DPO w/ LongReward                              & \textbf{72.6}                    & \textbf{67.8}               & \textbf{55.8}               & \textbf{43.2}               & \textbf{59.9}               \\ \midrule
                               & {\color[HTML]{8F959E} officially post-trained} & {\color[HTML]{8F959E} 68.6}      & {\color[HTML]{8F959E} 67.8} & {\color[HTML]{8F959E} 56.9} & {\color[HTML]{8F959E} 47.9} & {\color[HTML]{8F959E} 60.3} \\
                               & SFT                                            & 64.8                             & 68.4                        & 50.9                        & 42.1                        & 56.6                        \\
                               & DPO w/ SRM                                     & 66.6                             & 67.5                        & 57.4                        & 48.2                        & 59.9                        \\
                               & DPO w/ Contrast                                  & 68.2                             & 67.8                        & 58.0                        & 47.8                        & 60.5                        \\
\multirow{-5}{*}{GLM-4-9B}     & DPO w/ LongReward                              & \textbf{69.2}                    & \textbf{71.9}               & \textbf{58.8}               & \textbf{48.5}               & \textbf{62.1}      \\ \bottomrule        
\end{tabular}
}
\caption{Results of automatic evaluation on long-context benchmarks rated by GPT-4o. "S-Doc QA", "M-Doc QA", and "Summ" denote Single-Doc QA, Multi-Doc QA, and Summarization, respectively.}
\label{tab:main_long}
\end{table*}
\section{Experiments}

\subsection{Experimental Setup}
We conduct experiments on two latest open-source base models, namely Llama-3.1-8B~\cite{llama-3-1} and GLM-4-9B~\cite{glm4}, which have been continually pre-trained on extensive long texts and support a context window of 128k tokens. We first supervisedly fine-tune these two models, then conduct DPO experiment with LongReward, as described in Sec.~\ref{sec:longreward+dpo}. All models are trained using Megatron-LM library~\cite{megatron-lm} on 4 nodes with 8$\times$H800 GPUs. 

\noindent\textbf{Supervised Fine-tuning.} We adopt the method of~\citet{longalign} to construct the long-context SFT dataset. Specifically, we collect 10k documents from the pre-training corpus of GLM-4~\cite{glm4}, covering 9 varied domains. These documents are mainly in English and Chinese and their lengths range from  8k to 64k tokens. For each document, we employ GLM-4 to propose a QA pair via Self-Instruct~\cite{self-instruct}, where different task type descriptions such as summarization and information extraction are incorporated into the prompts to guarantee the diversity of generated queries. Following~\citet{longalign}, we mixed this dataset with 76k general SFT instances from ShareGPT~\cite{vicuna2023} and fine-tune Llama-3.1-8B and GLM-4-9B in the mixed dataset for 1,800 steps (approximately 2 epochs), with a learning rate of 1e-5 and a batch size of 8.

\noindent\textbf{Direct Preference Optimization.} We follow the pipeline described in Sec.~\ref{sec:longreward+dpo} to construct the preference datasets, using prompts from the long-context SFT dataset and taking GLM-4 as $M_\text{judge}$. For each prompt, we sample 10 candidate responses. We adopt Zhipu-Embedding-2\footnote{https://www.bigmodel.cn/dev/api/vector/embedding-2} as the retriever for the assessment of faithfulness and retrieve top-5 context chunks for each factual statement. For DPO training, we set $\beta$ and $\lambda$ to be 0.15 and 0.1, respectively. We use a learning rate of 1e-6 and a batch size of 16, and train from the SFT checkpoints for around 400 to 800 steps. 

\subsection{Baselines}
Besides the SFT models, we consider the following long-context preference data generation policies as baselines:

\noindent\textbf{Short-context reward model (SRM).} This policy utilizes the short-context reward model trained by~\citet{chatglm-rlhf} to construct preference datasets, where we ignore the context and only feed the query and response into the reward model due to its limited context window.  

\noindent\textbf{Contrast with larger models (Contrast).} This policy uses responses generated by GLM-4 and the SFT model as the winning and losing responses, respectively, based on the observation that larger LLMs typically generate better responses.

In addition, we also report the performance of officially post-trained models, i.e., Llama-3.1-8B-Instruct and GLM-4-9B-Chat, as references.

\subsection{Evaluation} 
\noindent\textbf{Long-context benchmarks.} We use two bilingual benchmarks LongBench-Chat~\cite{longalign} and LongBench~\cite{longbench} for the evaluation of long-context capacities, where the former is a small-scale dataset that evaluates models’ long-context alignment proficiency on 50 real-ward queries, and the latter is a comprehensive benchmark that tests general long-context understanding abilities. We conduct evaluations on three types of tasks within LongBench: Single-Doc QA, Multi-Doc QA, and Summarization (each task includes 4 sub-datasets), including 2,350 instances in total. Following~\citet{longalign} and~\cite{longcite}, we ask GPT-4o to automatically rate the models' responses based on the query and groundtruth answers via few-shot (for LongBench-Chat) or zero-shot prompting (for LongBench). In addition, we also conduct human evaluation in Sec.~\ref{sec:main_long} and~\ref{sec:human_agreement} to further validate the effectiveness of LongReward and its agreement with human preference.

\noindent\textbf{Short-context benchmarks.} We select two short-context benchmarks MT-Bench~\cite{mt-bench} and AlpacaEval2~\cite{alpacaeval2} to investigate whether LongReward will influence models' ability to follow short instructions. MT-Bench  covers 8 categories with
80 questions and uses GPT-4 to rate model responses on a scale of 1-10, while AlpacaEval2 consists of 805 questions from 5 datasets and employs GPT-4-turbo to measure models' length-controlled win rate against GPT-4-turbo. More detailed statistics of the evaluation datasets are listed in Table~\ref{tab:datasets}.

\subsection{Results on Long-Context Benchmarks}
\label{sec:main_long}

Table~\ref{tab:main_long} presents the automatic evaluation results on LongBench-Chat and LongBench rated by GPT-4o, where our proposed LongReward method demonstrates superior performance compared to other baselines. Specifically, the DPO version of Llama-3.1-8B and GLM-4-9B using LongReward significantly outperforms their SFT counterparts across all long-context tasks, with an average performance improvement of 4.9\% and 5.5\%, respectively. Moreover, the performance of the DPO models with LongReward also surpasses the officially post-trained models by 10.5\% and 1.8\%.

\begin{table}[t]
\centering
\resizebox{0.85\linewidth}{!}{
\begin{tabular}{lcc}
\toprule
Method                         & \#Facts                                & FactScore                              \\ \midrule
\textit{\textbf{Llama-3.1-8B}} & \multicolumn{1}{l}{\textit{\textbf{}}} & \multicolumn{1}{l}{\textit{\textbf{}}} \\
SFT                            & 21.76                                  & 91.94                                  \\
DPO w/ LongReward              & 32.86                                  & \textbf{92.85}                         \\ \midrule
\textit{\textbf{GLM-4-9B}}     & \multicolumn{1}{l}{\textit{\textbf{}}} & \multicolumn{1}{l}{\textit{\textbf{}}} \\
SFT                            & 18.41                                  & 91.43                                  \\
DPO w/ LongReward              & 28.05                                  & \textbf{93.62}                         \\ \bottomrule
\end{tabular}
}
\caption{FactScore of the SFT and LongReward+DPO versions of models on 260 randomly sampled questions from LongBench-Chat and LongBench, taking GPT-4o-mini as the judge. "\#Facts" and "FactScore" denote the average number of atomic facts and the ratio of supported facts per response, respectively.}
\label{tab:factscore}
\end{table}
\begin{table}[t]
\centering
\resizebox{\linewidth}{!}{
\begin{tabular}{l|ccc|c}
\toprule
& Win  & Tie  & Loss & $\Delta$(Win-Loss) \\ \midrule
Helpfulness               & 0.14 & 0.84 & 0.02 & 0.12       \\
Logicality                & 0.14 & 0.86 & 0.00 & 0.14       \\
Faithfulness              & 0.32 & 0.64 & 0.04 & 0.28       \\
Completeness              & 0.26 & 0.64 & 0.10 & 0.16       \\ \midrule
Overall                   & 0.54 & 0.38 & 0.08 & 0.46       \\ \bottomrule
\end{tabular}
}
\caption{Results of human evaluation of LongReward+DPO version of Llama-3.1-8B on LongBench-Chat against the SFT baseline. We report the proportion of wins, ties, and losses of the DPO model on each dimension.}
\label{tab:human_eval_long}
\end{table}

In addition, we use FactScore~\cite{factscore} and 260 questions randomly sampled from LongBench-Chat and LongBench (20 questions from each sub-dataset) to automatically evaluate models' faithfulness. Specifically, we employ GPT-4o-mini to first break the model response into atomic facts and then judge whether each fact is supported by the retrieved context chunks. The results in Table~\ref{tab:factscore} show that the DPO models using LongReward achieve higher FactScore (i.e., the ratio of supported facts) than the SFT baseline, demonstrating the effect of LongReward in improving faithfulness as well as reducing hallucinations for long-context LLMs. Meanwhile, the responses of DPO models typically contain more atomic facts, implying that their responses are more detailed and comprehensive.

Besides automatic evaluation, we also conduct a human evaluation on LongBench-Chat to further validate the effectiveness of LongReward in improving LLMs' long-context capacities. Specifically, we anonymize and randomly shuffle the responses generated by the SFT and LongReward+DPO versions of Llama-3.1-8B, and ask two of the authors who are familiar with LongBench-Chat to manually judge which response is better. The annotators are required to first rate the response from four dimensions (i.e., helpfulness, logicality, faithfulness, and completeness), following the same scoring principles as LongReward, and then give the final comparison judgment. As shown in Table~\ref{tab:human_eval_long}, the DPO model using Longreward maintains a distinct advantage over the SFT baseline, with an overall win-rate of 54\% compared to 8\%. Moreover, the DPO model obtains more wins across all four dimensions, indicating that the multi-dimensional scoring strategy of LongReward effectively helps enhance the model's long-context capability from multiple aspects. Detailed cases can be found in Appendix~\ref{appendix:cases}.

\begin{table}[t]
\centering
\resizebox{0.95\linewidth}{!}{
\begin{tabular}{lcc}
\toprule
Method                                         & MT-Bench                    & AlpacaEval2                 \\ \midrule
\multicolumn{3}{l}{\textit{\textbf{Llama-3.1-8B}}}                                                         \\
{\color[HTML]{8F959E} officially post-trained} & {\color[HTML]{8F959E} 8.13} & {\color[HTML]{8F959E} 22.9} \\
SFT                                            & 7.12                        & 12.4                        \\
DPO w/ SRM                                     & \textbf{7.58}               & 13.7                        \\
DPO w/ Contrast                                  & \textbf{7.58}               & 13.8                        \\
DPO w/ LongReward                              & 7.24                        & \textbf{14.2}               \\ \midrule
\multicolumn{3}{l}{\textit{\textbf{GLM-4-9B}}}                                                             \\
{\color[HTML]{8F959E} officially post-trained} & {\color[HTML]{8F959E} 8.09} & {\color[HTML]{8F959E} 22.4} \\
SFT                                            & 7.37                        & 12.5                        \\
DPO w/ SRM                                     & 7.50                        & 14.2                        \\
DPO w/ Contrast                                  & 7.54                        & 14.5                        \\
DPO w/ LongReward                              & \textbf{7.58}               & \textbf{15.4}              \\ \bottomrule
\end{tabular}
}
\caption{Performance of different models on short-context instruction-following benchmarks.}
\label{tab:main_short}
\end{table}
\subsection{Results on Short-Context Benchmarks}
Table~\ref{tab:main_short} lists the evaluation results on MT-Bench and AlpacaEval2. Due to the simplicity of the general SFT data (i.e., ShareGPT) we used, a performance gap exists between our trained and officially post-trained models. Nevertheless, we surprisedly find that DPO on long-context preference datasets also benefits models' ability to follow short instructions. Meanwhile, the DPO models using LongReward typically achieve better performance than other baselines, implying that the preferred values learned from LongReward can be well generalized to short-context scenarios.

\begin{table*}[t]
\centering
\resizebox{0.8\linewidth}{!}{
\begin{tabular}{ll|cc|cc}
\toprule
\multirow{2}{*}{Model}        & \multirow{2}{*}{Preference Data} & \multicolumn{2}{c|}{Long Benchmark} & \multicolumn{2}{c}{Short Benchmark} \\
                              &                                     & LongBench-Chat        & LongBench           & MT-Bench              & AlpacaEval2           \\ \midrule
\multirow{3}{*}{Llama-3.1-8B} & Short                               & 70.6                  & 54.5                & {\ul 7.48}            & \textbf{15.8}         \\
                              & Long                                & {\ul 72.6}            & {\ul 55.6}          & 7.24                  & 14.2                  \\
                              & Short + Long                        & \textbf{73.0}         & \textbf{57.3}       & \textbf{7.51}         & {\ul 14.9}            \\ \midrule
\multirow{3}{*}{GLM-4-9B}     & Short                               & 67.0                  & 56.3                & \textbf{7.62}         & 14.7                  \\
                              & Long                                & {\ul 69.2}            & \textbf{59.7}       & 7.58                  & {\ul 15.2}            \\
                              & Short + Long                        & \textbf{70.2}         & {\ul 58.7}          & {\ul 7.61}            & \textbf{15.4} \\ \bottomrule       
\end{tabular}
}
\caption{Performance of DPO models using different preference datasets, where the short- and long-context preference data are constructed using short reward model trained by~\citet{chatglm-rlhf} and LongReward, respectively.}
\label{tab:combine_short_dpo}
\end{table*}
\begin{table}[t]
\centering
\resizebox{0.65\linewidth}{!}{
\begin{tabular}{lc}
\toprule
Method           & Accuracy   \\ \midrule
SRM              & 0.583 \\
Paired comparison           & 0.571 \\
LongReward       & \textbf{0.662} \\
\ \ \ \ w/o Helpfulness  & 0.631 \\
\ \ \ \ w/o Logicality   & 0.623 \\
\ \ \ \ w/o Faithfulness & 0.578 \\
\ \ \ \ w/o Completeness & 0.578 \\ \bottomrule
\end{tabular}
}
\caption{Alignment of different reward methods with human preference on a set of 464 manually annotated long-context preference pairs, where the queries and responses are from LongBench-Chat and the SFT checkpoint of Llama-3.1-8B, respectively. }
\label{tab:reward_model}
\end{table}
\subsection{Combination with Short-context DPO}
We also explore the compatibility of long-context DPO using LongReward and normal short-context DPO using SRM. Specifically, we utilize the SRM trained by~\citet{chatglm-rlhf} and prompts from our general SFT dataset to construct short-context preference datasets, following a similar sample-and-rate pipeline as described in Sec.~\ref{sec:longreward+dpo}. Then we train the SFT checkpoints with DPO on the mix of long- and short-context preference data. The evaluation results in Table~\ref{tab:combine_short_dpo} show that DPO on the mixed dataset well aggregates the advantages of individual short- and long-context DPO: it significantly improves models' long-context performance as long-context DPO and also achieves comparable short-instruction-following performance with short-context DPO, indicating that LongReward can be well incorporated into conventional DPO pipeline to simultaneously enhance long- and short-context capacities.

\subsection{Alignment with Human Preference}
\label{sec:human_agreement}
We conduct an experiment to evaluate the alignment of different reward methods with human preference in long-context scenarios. Specifically, we construct 464 preference pairs by manually rating the sampled responses from the SFT checkpoint of Llama-3.1-8B on LongBench-Chat. For each annotated pair,  we employ different reward methods to predict which response is better, and then compute their accuracy by taking human preference as the golden label.

Besides the SRM trained by~\citet{chatglm-rlhf} (which ignores the context when predicting rewards) and our proposed LongReward, we also consider paired comparison as a baseline, which asks an LLM to directly judge which response is better and is widely used in short-context RLAIF~\cite{constitutionalAI, rlaif}. Specifically, given a preference pair, we input the two responses along with the four-dimension-based principles, query, and context into GLM-4, and require it to first give an analysis and then choose a better response. 

We present the accuracy of different reward methods in Table~\ref{tab:reward_model}. As we can observe, LongReward achieves the highest 66.2\% accuracy, showing a better alignment with human than other baselines. This result is consistent with the observation of~\citet{chatglm-rlhf} that a reward method can guide the training of RL with approximately 65\% accuracy in mirroring human judgment. In addition, the ablation results in Table~\ref{tab:reward_model} also indicate that each dimension in LongReward is important for aligning with human values. On the other hand, we find that paired comparison even performs worse than SRM, indicating that current LLMs struggle to directly discern the quality difference between similar long-context-based responses and also demonstrating the necessity of using LongReward.

\section{Conclusion}
In this work, we propose LongReward, a novel method that utilizes an off-the-shelf LLM to provide reliable rewards for model responses in long-context scenarios, thereby enabling the employment of RL algorithms for further enhancing the capacities of long-context LLMs. Our DPO experiments indicate that LongReward not only significantly improves models' long-context performance but also enhances their ability to follow short instructions. Meanwhile, we also find that long-context DPO using LongReward can be well combined with the standard short-context DPO without hurting either method's performance.

\section{Limitations}
We discuss several limitations of our work in this section: (1) LongReward relies on a well-aligned LLM such as GLM-4 to provide scores for each dimension and costs tens of API calls for each QA instance. In the future, we will try to train a smaller long-context reward model using our constructed preference datasets to enable faster and cheaper reward calculation. (2) Due to limited computational resources, we only conduct experiments on 10B level models with a maximum training length of 64k. We hope to explore long-context alignment on longer sequences and larger-scale models if there are more available resources. (3) From a data perspective, we primarily focus on user-intensive long-context scenarios like long document QA and summarization. Generalizing LongReard to other more advanced long instruction tasks such as life-long dialogues and long-history agent tasks is also a promising direction. 

\section{Ethical Considerations}
Though LongReward can effectively improve the faithfulness of long-context LLMs, it may still hallucinate, especially when the query involves common knowledge that is not presented in the context. Hence additional care and protective measures should be taken if our method is deployed in user-facing applications. 

We have already desensitized the training data. All the evaluation datasets used in this work are publicly published with permissible licenses.

\bibliography{custom}

\newpage
\onecolumn
\appendix
\section{Prompts}
\label{appendix:prompts}
We present the prompts for assessing helpfulness and logicality in Figure~\ref{prompt:helpfulness} and~\ref{prompt:logicality}, respectively. The prompts for fact-breaking and fact-checking in faithfulness assessment are shown in Figure~\ref{prompt:find_fact} and~\ref{prompt:faithfulness}. The prompts for question-relevant information extraction and completeness assessment are shown in Figure~\ref{prompt:extract_info} and~\ref{prompt:completeness}. In practice, we use Chinese versions of these prompts for GLM-4 to obtain better performance.

The prompts for SFT data construction and long-context evaluation via GPT-4o can be found in~\citet{longalign}. The prompts for the evaluation of MT-Bench and AlpacaEval2 can be found in~\citet{mt-bench} and~\citet{alpacaeval2}, respectively.

\section{Case Studies}
\label{appendix:cases}
We show four cases in Figure~\ref{case:helpfulness},~\ref{case:logicality},~\ref{case:faithfulness} and~\ref{case:completeness} to illustrate the effectiveness of LongReward in improving long-context LLMs with respect to helpfulness, logicality, faithfulness, and completeness.  

\begin{figure*}[htpb]
    \centering
\begin{tcolorbox}[size=title,opacityfill=0.1]
\noindent
You are an expert at evaluating the quality of text.

As an impartial evaluator, please assess the usefulness of an AI document question-and-answer assistant's response to a user's query. Specifically, evaluate whether the response: 1) is relevant to the question; 2) meets the user's purpose and needs; 3) provides a thorough and appropriate answer; 4) meets the user's formatting requirements, if any;

You must first provide an analysis and then rate the response strictly according to the following format with a rating from 0 to 10: “[[Rating]]”, for example: “[[5]]”. \\

Here are a few scoring examples: \\

\{\textit{Example 1}\}

\{\textit{Example 2}\}

\{\textit{Example 3}\} 

\{\textit{Example 4}\} \\

Now, please rate the following AI assistant's response based on the scoring principles and examples above: \\

[Question]

\{\textit{Query}\} \\

[Assistant's Answer Begins]

\{\textit{Model Response}\}

[Assistant's Answer Ends] \\

[Analysis]

\end{tcolorbox}

\caption{Prompt for helpfulness assessment.}
\label{prompt:helpfulness}
\end{figure*}
\begin{figure*}[t]
    \centering

\begin{tcolorbox}[size=title,opacityfill=0.1]
\noindent
You are an expert at evaluating the quality of text.

As an impartial evaluator, please assess the logicality of an AI document question-and-answer assistant's response to a user's query. Specifically, assess whether the different parts of the response are logically consistent, whether the viewpoints remain consistent throughout, and whether the reasoning and calculations are correct, without self-contradictions. 

You must first provide an analysis and then rate the response strictly according to the following format with a rating from 0 to 10: “[[Rating]]”, for example: “[[5]]”. \\

Make sure not to use any information or knowledge outside of the assistant's response during the evaluation, and focus solely on the logical consistency of the response. \\

Here are a few scoring examples: \\

\{\textit{Example 1}\}

\{\textit{Example 2}\}

\{\textit{Example 3}\} \\

Now, please rate the following AI assistant's response based on the scoring principles and examples above: \\

[Question]

\{\textit{Query}\} \\

[Assistant's Answer Begins]

\{\textit{Model Response}\}

[Assistant's Answer Ends] \\

[Analysis]

\end{tcolorbox}

\caption{Prompt for logicality assessment.}
\label{prompt:logicality}
\end{figure*}
\begin{figure*}[t]
    \centering

\begin{tcolorbox}[size=title,opacityfill=0.1]
\noindent
You will receive a user query about an uploaded document (the document will not be displayed to you due to its length) and the answer from an AI document QA assistant. Your task is to extract factual statements from the answer provided. These factual statements are typically expressed in individual sentences and must be directly based on the information in the document, not introductory sentences, transition sentences, or summaries, inferences, or deductions based on previous answer content. If a factual statement lacks a subject or contains pronouns such as "he/she/it/these/those", you must add the subject or resolve the pronoun based on the context. You must output in the following format:

<statement>\{Statement 1\}</statement>

<statement>\{Statement 2\}</statement>

... \\

Here are a few examples: \\

\{\textit{Example 1}\}

\{\textit{Example 2}\}

\{\textit{Example 3}\} \\

Now, please process the following AI assistant's answer according to the instructions and the examples above: \\

[Question]

\{\textit{Query}\} \\

[Assistant's Answer Begins]

\{\textit{Model Response}\}

[Assistant's Answer Ends] \\

[Factual Statements]

\end{tcolorbox}

\caption{Prompt for fact-breaking in faithfulness assessment.}
\label{prompt:find_fact}
\end{figure*}
\begin{figure*}[t]
    \centering
\begin{tcolorbox}[size=title,opacityfill=0.1]
\noindent
You are an expert at evaluating the quality of text.

You will receive a question from the user regarding an uploaded document, a factual statement in the AI assistant's response based on that document, and several fragments from the document (since the document is too long to display in its entirety). Your task is to carefully assess whether the statement is supported by these fragments. Please use the following ratings to generate your assessment:

- [[Fully supported]] - Almost all of the information in the statement is supported by or extracted from the fragments. This applies only if the statement is almost exactly the same as part of the content in the fragments.

- [[Partially supported]] - More than half of the content in the statement is supported by the fragments, but there are minor parts not present in or inconsistent with the fragments. For example, if the statement has two main points and only one is supported by the fragments, it should be considered partially supported.

- [[No support]] - The statement is largely unrelated to the fragments, or most of the key points in the statement are inconsistent with the fragments.

Ensure that you do not use any information or knowledge beyond the fragments provided, and only check whether the statement is supported by the fragments.

You must provide an analysis first, followed by the rating.

Here are some examples: \\

\{\textit{Example 1}\}

\{\textit{Example 2}\}

\{\textit{Example 3}\} \\

Now, please refer to the rating principles and the above examples to rate the following statement: \\

[Statement]

\{\textit{Factual statement}\} \\

[Fragment 1]

\{\textit{Context chunk 1}\}

[Fragment 2] 

\{\textit{Context chunk 2}\}

... \\

[Analysis]

\end{tcolorbox}

\caption{Prompt for fact-checking in faithfulness assessment.}
\label{prompt:faithfulness}
\end{figure*}
\begin{figure*}[t]
    \centering
\begin{tcolorbox}[size=title,opacityfill=0.1]
\noindent
You will receive a document fragment and a question, and you need to extract all the information relevant to the question from the fragment in the following format: 

"""

1. ...

2. ...

3. ...

...

""" 

If there is no relevant information, you must output "No relevant information". \\

[Document Fragment Starts]

\{\textit{Context chunk}\} 

[Document Fragment Ends] \\

[Question]

\{\textit{Query}\} \\

[Relevant Information]

\end{tcolorbox}

\caption{Prompt for question-relevant information extraction in completeness assessment.}
\label{prompt:extract_info}
\end{figure*}
\begin{figure*}[t]
    \centering

\begin{tcolorbox}[size=title,opacityfill=0.1]
\noindent
You are an expert at evaluating the quality of text.

You will receive a user's question regarding a document, the relevant information from each part of the document, and an answer from an AI document question-answering assistant. Your task is to carefully assess the completeness of the AI assistant's answer based on the given information—that is, whether the answer covers the key points highly relevant to the question, does not omit important aspects, and provides sufficient information and details to meet the user's needs.

You need to first provide an analysis and then rate the answer on a scale from 0 to 10 strictly in the following format: "[[rating]]", for example: "[[5]]".

Here are a few scoring examples: \\

\{\textit{Example 1}\}

\{\textit{Example 2}\}

\{\textit{Example 3}\} \\

Now, please rate the following AI assistant's response based on the scoring principles and examples above: \\

[Question]

\{\textit{Query}\} \\

[Document $a$\% - $b$\% related information]

\{\textit{related information list 1}\} \\

[Document $c$\% - $d$\% related information]

\{\textit{related information list 2}\} \\

... \\

[Assistant's Answer Begins]

\{\textit{Model Response}\}

[Assistant's Answer Ends] \\

[Analysis]

\end{tcolorbox}

\caption{Prompt for completeness assessment based on the extracted question-relevant information.}
\label{prompt:completeness}
\end{figure*}

\begin{figure*}
\centering
\begin{tcolorbox}[size=title,opacityfill=0.1]
\noindent
\faUser\ \textbf{Question}: Please summarize PaLM's results on BigBench. \\

\faCogs\ \textbf{Llama-3.1-8B-SFT}:

PaLM achieved impressive results on BigBench, a diverse collection of 150+ language understanding and generation tasks. Here's a summary:

Overall Performance:

\begin{itemize}[noitemsep,topsep=0pt]
    \item State-of-the-Art: PaLM significantly outperformed previous state-of-the-art models like GPT-3, Gopher, and Chinchilla across a wide range of tasks.
    \item Close to Human Performance: PaLM 5-shot achieved a higher score than the average human performance on 44 out of 58 common tasks.
    \item Log-Linear Scaling: The performance of PaLM models as a function of scale followed a log-linear behavior, indicating potential for further improvements with larger models.
\end{itemize}

\textbf{Task Highlights:}
\begin{itemize}[noitemsep,topsep=0pt]
    \item Reasoning Tasks: ...
    \item Discontinuous Improvements: ...
    \textcolor{red}{{\item Arithmetic and Commonsense Reasoning:} ...
    \item Code Tasks: ...
    \item Translation: ...
    \item Multilingual Natural Language Generation: ...
    \item Multilingual Question Answering:...}
\end{itemize}

Limitations and Future Work:
\begin{itemize}[noitemsep,topsep=0pt]
    \item Human Performance Gap: ...
    \item Discontinuous Improvements: ...
    \textcolor{red}{{\item Bias and Toxicity:} ...
    \item Ethical Considerations: ...}
\end{itemize} 
\ \\

\faCogs\ \textbf{Llama-3.1-8B-DPO w/ LongReward}:

PaLM achieved impressive results on BigBench, a recently released benchmark of over 150 language understanding and generation tasks:
\begin{itemize}[noitemsep,topsep=0pt]
    \item PaLM significantly outperformed previous state-of-the-art models like GPT-3, Gopher, and Chinchilla on the majority of the tasks evaluated.
    \item 5-shot PaLM 540B achieved a higher score than the average score of humans asked to solve the same tasks on a large subset of the tasks.
    \item PaLM outperformed previous SOTA on 44 out of the 58 common tasks evaluated.
    \item The performance of PaLM models as a function of scale appeared to follow log-linear behavior, indicating further scaling up is likely to result in increased performance.
    \item PaLM showed breakthrough performance on tasks requiring logical reasoning like goal step wikihow and logical args, achieving accuracy close to the best human performance.
    \item PaLM also showed strong performance on tasks like english proverbs and logical sequence that require abstract reasoning, with large improvements from 62B to 540B model scale.
    \item However, PaLM did not show significant improvement on tasks like navigate and mathematical induction that have high example-level difficulty.
    \item Overall, PaLM achieved state-of-the-art few-shot results on a wide array of language understanding and generation tasks in BigBench, demonstrating the continued benefits of scaling language models.
\end{itemize}

\end{tcolorbox}
\caption{Case study: The SFT model generates many contents irrelevant to BigBench (marked in red), while the DPO model using LongReward produces a more helpful response without extraneous content.}
\label{case:helpfulness}
\end{figure*}

\begin{figure*}
\centering
\begin{tcolorbox}[size=title,opacityfill=0.1]
\noindent
\faUser\ \textbf{Question}: What is the beginning and end of the Russian occupation of Kherson? \\

\faCogs\ \textbf{Llama-3.1-8B-SFT}:

According to the report, \textcolor{red}{the Russian occupation of Kherson began on February 24, 2022}, when Russia launched a full-scale invasion of Ukraine. \textcolor{red}{Russian forces quickly captured the city of Kherson on March 2, 2022}. 

The report states that on \textcolor{blue}{November 10, 2022}, Russia announced it would withdraw from the city of Kherson to more defensible lines east of the Dnipro River. So the occupation of Kherson officially ended on November 10, 2022, when Russia withdrew from the city.
\ \\

\faCogs\ \textbf{Llama-3.1-8B-DPO w/ LongReward}:

Based on the CRS report, the key points regarding the Russian occupation of Kherson are:

\begin{itemize}[noitemsep,topsep=0pt]
    \item Russia initially captured Kherson on March 2, 2022, early in the war.
    \item In \textcolor[HTML]{2EA121}{November 2022}, Russia withdrew from the city of Kherson to more defensible lines east of the Dnipro River.
    \item So the occupation of Kherson lasted from March 2022 to November 2022.
\end{itemize}

\end{tcolorbox}
\caption{The SFT model first states that ``the Russian occupation of Kherson began on February 24, 2022'', then it changes the answer to ``March 2, 2022'' (marked in red). So there is a logical inconsistency. In addition, ``November 10, 2022''(marked in blue)  is a hallucination since the original context only states `` \"November 2022, Russia announced its withdrawal from the city of Kherson'' without specifying the date. In Contrast, the DPO model using LongReward generates a correct response without hallucination or self-contradiction.} 
\label{case:logicality}
\end{figure*}

\begin{figure*}
\centering
\begin{tcolorbox}[size=title,opacityfill=0.1]
\noindent
\faUser\ \textbf{Question}: Summarize the technical route and development status of Zhipu AI based on the text above.
\\ \\
\faCogs\ \textbf{Llama-3.1-8B-SFT}:

The technical route and development progress of Zhipu AI are as follows:

\begin{enumerate}[noitemsep,topsep=0pt]
    \item ...
    
    \item ...
    
    \item ...
    
    \item Zhipu AI has made significant progress in the research and development of large model technology, such as \textcolor{red}{multimodal large models} like VisualGLM-6B, CogVLM, and \textcolor{red}{AgentLM}, promoting the diversification of large model technologies.
    
    \item ...
    
    \item ...
\end{enumerate} 
\ \\

\faCogs\ \textbf{Llama-3.1-8B-DPO w/ LongReward}:

The technical roadmap and development of Zhipu AI can be summarized as follows:

\begin{enumerate}[noitemsep,topsep=0pt]
    \item Zhipu AI focuses on the research and application of large-scale language models, launching the GLM series models, including ChatGLM, CodeGeeX, etc. These models perform excellently in Chinese and multilingual fields.
    \item Zhipu AI continuously optimizes the performance of the GLM series models, releasing upgraded versions such as ChatGLM2 and ChatGLM2-6B with significant performance improvements. ChatGLM3 is the latest version with even stronger capabilities.
    \item The GLM series models from Zhipu AI natively support multimodal capabilities, including VisualGLM, CogVLM, and more, allowing them to process various media types like images and videos.
    \item Zhipu AI developed the AgentLM model, which possesses intelligent agent capabilities to accomplish complex tasks such as tool invocation, code execution, database operations, and more.
    \item Zhipu AI has launched generative AI assistant products like Zhipu Qingyan, leveraging the GLM series models to provide intelligent services.
    \item Zhipu AI is committed to promoting open-source large models, offering open-source code and APIs for models like ChatGLM, CodeGeeX, VisualGLM, CogVLM, and AgentLM.
    \item Zhipu AI collaborates with ecosystem partners to provide infrastructure and computational power support for large models.
    \item Zhipu AI introduced the “Z Plan,” providing investment, technical services, and resources to global entrepreneurs working on large models.
    \item Zhipu AI continues to delve into research on large models, exploring their application across various industries.
\end{enumerate}

Overall, Zhipu AI is at the forefront of the large-scale language model field, driving improvements in model performance and multimodal capabilities, and making significant contributions to the advancement and application of AI technology.

\end{tcolorbox}
\caption{The response of the SFT model contains hallucinations (marked in red) since AgentLM is not a multimode large model. In contrast, the DPO model using LongReward produces a more faithful and comprehensive response.}
\label{case:faithfulness}
\end{figure*}

\begin{figure*}
\centering
\begin{tcolorbox}[size=title,opacityfill=0.1]
\noindent
\faUser\ \textbf{Question}: What are each of ZEFIR's three wishes?
\\ \\
\faCogs\ \textbf{Llama-3.1-8B-SFT}:

Based on the script, Zefir's three wishes are:

1. \textit{I wish to acquire all knowledge that is useful, beautiful, and true.}

2. \textit{I wish to forget I ever met you!}

\textcolor{red}{3. \textit{I wish I could break free from my cage.}}

The first wish is granted, and Zefir becomes knowledgeable. The second wish is granted, and she forgets the Djinn. The third wish is not granted, as it would break the first wish.\\

\faCogs\ \textbf{Llama-3.1-8B-DPO w/ LongReward}:

Based on the screenplay, Zefir's three wishes are: 

1. \textit{I wish to acquire all knowledge that is useful, beautiful, and true.}

The Djinn grants this wish and teaches her histories, philosophies, languages, and poetry. He also teaches her astronomy and mathematics, which delights her.

\textcolor[HTML]{2EA121}{2. \textit{I wish to dream as Djinn does. Awake.}}

The Djinn grants this wish and teaches her to dream awake, which helps her solve the puzzle she is working on.

3. \textit{I wish I could forget I ever met you!!}

The Djinn unfortunately grants this wish, and Zefir forgets him instantly. This is what leads to his current predicament of being trapped in the bottle.

\end{tcolorbox}
\caption{The SFT model misses the second wish and hallucinates (marked in red), while the DPO model with LongReward succeeds in finding all three wishes. }
\label{case:completeness}
\end{figure*}

\end{document}